\documentclass{article}

 \usepackage[preprint]{neurips_2026}

\usepackage[utf8]{inputenc} 
\usepackage[T1]{fontenc}    
\usepackage{hyperref}       
\usepackage{url}            
\usepackage{booktabs}       
\usepackage{amsfonts}       
\usepackage{nicefrac}       
\usepackage{listings}
\usepackage{microtype}      
\usepackage{xcolor}         
\usepackage{amsmath}
\newcommand{\method}{\mbox{EHR-FM}}
\usepackage{subcaption}
\usepackage[most]{tcolorbox}
\usepackage{graphicx}
\usepackage{multirow}
\usepackage{wrapfig}
\usepackage{caption}

\newtcolorbox[list inside=prompt,auto counter,number within=section]{prompt}[1][]{
    colbacktitle=black!60,
    coltitle=white,
    fontupper=\footnotesize,
    boxsep=5pt,
    breakable,
    enhanced,
    left=0pt,
    right=0pt,
    top=0pt,
    bottom=0pt,
    boxrule=1pt,
    title=#1,
}

\definecolor{midnightgreen}{rgb}{0.0, 0.29, 0.33}

\title{Scaling Electronic Health Record Foundation Models for Population Health Management}

\author{%
    \textbf{Liwen Sun}$^1$ \quad
    \textbf{Hao-Ren Yao}$^{4}$ \quad
    \textbf{Ophir Frieder}$^{3,4}$ \quad
    \textbf{Xiang Qian}$^{2,4}$ \quad
    \textbf{Chenyan Xiong}$^{1,4}$ \quad \\
    $^{1}$ Carnegie Mellon University \quad
    $^{2}$ Stanford University \quad
    $^{3}$ Georgetown University \quad
    $^{4}$ Xlue\\
}
\begin{document}

\maketitle

\begin{abstract}
Population health management requires scalable methods to identify individuals at risk of chronic diseases such as cardiovascular conditions and cancer, yet existing approaches rely on fragmented data and resource-intensive screening. We present \method{}, an Electronic Health Record Foundation Model that performs large-scale chronic disease prediction using cross-site longitudinal medical records. We pretrain \method{} on billions of medical events from over 5 million patients across Taiwan and the United States, leveraging a unified code alignment framework to address cross-system heterogeneity, and characterize its scaling behavior via IsoFLOP analysis, training compute-optimal models up to 2.4B parameters. Across 11 chronic disease prediction tasks, \method{} demonstrates strong scaling and generalization, outperforming tree-based models and both general and biomedical language models, achieving over 40\% and 70\% sensitivity at 99\% specificity in U.S. and Taiwan cohorts, respectively. On the EHRShot benchmark, \method{} surpasses prior EHR foundation models trained on in-site data despite substantial distribution shifts, highlighting strong few-shot generalization. Finally, we show that aligned cross-system data provides more effective pretraining signal than duplicating single-site data under data-limited settings, underscoring the importance of alignment for scalable healthcare modeling. Our analysis demonstrates the robustness of EHR-FM in various patient distributions and the benefits of operating in the ICD code space. The code will be open-sourced.

\end{abstract}
\section{Introduction}
Population health management is a key strategy for improving health outcomes by enabling the systematic identification and monitoring of individuals at elevated risk for chronic diseases and adverse events \citep{mcnamara2024phm}. Proactive risk stratification and coordinated care have been shown to reduce morbidity and improve outcomes in conditions such as cardiovascular disease, stroke, and cancer \citep{bredehorst2024ascvd,kim2024characteristics,gynnild2021stroke}. Early identification of high-risk populations allows for timely preventive interventions, improved disease control, and reduced healthcare utilization \citep{rajkomar2018scalable,adini2019early}. Accordingly, longitudinal monitoring and integrated care delivery have become essential components of modern healthcare systems, shifting the focus from reactive treatment to prevention and early intervention \citep{gyawali2024treatment, rubinstein2024cancer}.

Despite its importance, population health management remains inadequately implemented in many settings \citep{wang2023barriers}. Healthcare delivery is often fragmented, with limited coordination across providers and incomplete integration of patient data, leading to missed opportunities for early intervention \citep{kern2024fragmentation,ghannam2025fragmentation}. Preventive strategies frequently rely on resource-intensive or episodic care models, which are difficult to scale and may not effectively reach high-risk or underserved populations \citep{rohatgi2020peer, washington2020health}. As a result, many individuals with emerging or asymptomatic conditions are not identified until disease progression, leading to worse outcomes and higher healthcare costs \citep{blackford2024pancreatic}. These limitations highlight the urgent need for more effective and scalable population health management approaches.

We present  \method{}, Electronic Health Record Foundation Model, a scalable approach for population health management that leverages cross-site longitudinal electronic medical records to identify individuals at elevated risk for major chronic diseases, including cardiovascular disease and cancer. EHR-FM is pretrained on large-scale longitudinal EHR data and learns generalizable patterns of disease progression directly from structured medical codes (e.g., ICD) through next-code prediction, enabling it to capture clinically meaningful signals embedded in routine care. 

A major challenge in leveraging EHR at scale is the heterogeneity of coding systems across countries and institutions, where differences in ICD versions, local code sets, and clinical practices hinder unified representation and cross-site generalization \citep{harmonize2026,kirchler2026transferability}. In this work, we integrate data from over 3 million patients in Taiwan and 2 million patients from multiple U.S. states.
To address this challenge, we utilize a unified code alignment framework that standardizes heterogeneous medical codes into a shared space, enabling generalized modeling across diverse healthcare systems while preserving clinically meaningful information.

With billions of medical events available, we further examine how compute budget (FLOPs), model size, and pretraining tokens affect performance, establishing an EHR foundation model scaling law and pretrain compute-optimal EHR-FM models up to 2.4B parameters in cross-site settings. When fine-tuned and evaluated on 11 chronic disease prediction tasks across cross-site data from the U.S. and Taiwan, \method{} exhibits strong scaling and generalization behavior and outperforms feature-based tree models as well as both general and medical large language models. It achieves strong predictive performance, with over 40\% and 70\% sensitivity at 99\% specificity in the U.S. and Taiwan cohorts, respectively, and reaches 30\% and 70\% AUPRC on average across tasks.

Moreover, on new diagnosis assignment tasks from the public EHRSHOT benchmark \citep{wornow2023ehrshot}, \method{} outperforms prior state-of-the-art foundation model, CLMBR, trained on millions of in-site patient records, despite being trained on data from substantially different populations, disease prevalence, and coding systems, which highlights its strong few-shot generalization capability. We also show that aligned cross-site EHR data provides more effective pretraining signal than duplicating single-site data under data-limited settings, leading to improved performance and underscoring the value of cross-site data integration. Finally, our contributions can be summarized as follows:

\begin{itemize}

\item We characterize scaling behavior for EHR foundation models via IsoFLOP analysis, deriving power-law relationships between compute, model size, and training tokens in cross-site medical settings to reveal the balance between model and data scale.

\item We demonstrate that scaling EHR-FMs improves performance across 11 chronic disease prediction tasks in cross-site settings, and also achieves strong few-shot generalization on the public EHRShot benchmark.

\item We show that aligned cross-system EHR data provides more effective pretraining signal at scale: under data-limited settings, duplicating single-site data yields limited gains, whereas incorporating aligned cross-system data improves performance.
\end{itemize}

To the best of our knowledge, we present the first scaled and largest EHR foundation model for population health management in a cross-site setting spanning countries and institutes. By leveraging longitudinal medical records from diverse healthcare systems, EHR-FM enables low-cost, large-scale risk stratification, facilitating early identification of high-risk populations and timely preventive interventions while addressing limitations of fragmented data and resource-intensive screening.

\section{Related Work}
Early detection is critical to chronic disease prevention in population health management, spanning cancer and cardiovascular disease. Cancer screening improves outcomes~\citep{kim2024characteristics, kukhareva2024lung, thiele2024screening}, and recent advances in AI-powered medical imaging (e.g., CT-based multimodal models) have further improved screening accuracy~\citep{cao2023large, chen2023pancreatic, wang2024pathology, xu2024whole, sun2024fact}. However, imaging remains resource-intensive and inaccessible for under-served populations, limiting scalability~\citep{elmohr2024social, waite2021narrowing}. In parallel, clinical practice for cardiovascular diseases relies on guideline-driven risk calculators~\citep{anderson2024ascvd}. Yet these tools use limited static measurements and show calibration issues and inconsistent performance across populations, limiting their effectiveness for dynamic risk prediction and large-scale population stratification.

Using electronic health records (EHR) to assess chronic disease risk is a promising path to improve detection effectiveness and efficiency by identifying patients with high risk for healthcare professionals to make informed decisions~\citep{lee202110,lee2022prediction}. 
Recent approaches have made attempts using feature-based machine learning models  to detect chronic disease risks and have shown the possibility of capturing risk signals in EHR~\citep{peduzzi2024explainable, placido2023deep}.

Inspired by the success of pretrained large language models on medical corpora~\citep{singhal2025toward,clusmann2023future,liu2025generalist},
researchers has also pretrained foundation models on structured EHR data~\citep{renc2024ethos,mcdermott2023eventstreamgptdata}, and explored their utilities in various clinical tasks~\citep{savcisens2024using, yang2023transformehr}.
A challenge for EHR foundation models is that many EHR datasets only cover a slice of patients' healthcare records~\citep{faltys2021hirid, johnson2016mimic, pollard2018eicu}, e.g., MIMIC-IV mainly focuses on emergency departments and ICU encounters~\citep{johnson2016mimic}. Foundation models pretrained on EHR slices are more effective in prediction tasks such as emergency department triage~\citep{suned2024} and ICU readmission risk~\citep{jiang2023health}.

Many tasks, like cancer risk prediction, require longitudinal EHR data to capture full patient history. Foresight uses two decades of data to pretrain transformers for biomedical forecasting~\citep{kraljevic2022foresight}. MOTOR is pretrained on  2.7M  records from 2014-2022 for time-to-event tasks to predict diagnosis time~\citep{steinberg2023motor}. CLMBR is pretrained on 2.6M records over three decades, with 6.7K records released in the EHRSHOT benchmark for 15 few-shot prediction tasks~\citep{wornow2023ehrshot}.

Although prior work on EHR foundation models has explored longitudinal disease modeling and scaling behavior~\citep{waxler2025generative,zhang2025exploring}, they are typically limited to single-site settings and do not account for non-trivial heterogeneity in coding systems across countries and institutions. Consequently, cross-site generalization remains underexplored. We address this gap by studying the scaling of EHR foundation models in cross-site settings, enabled by aligning coding systems across diverse healthcare systems.


\section{Cross-site Data curation}
Here, we describe our data curation across sites with unified code alignment.
\subsection{Medical Code Alignment}
Heterogeneous healthcare systems use different coding schemes, hindering cross-system scaling of EHR foundation models. The same clinical concept may appear as distinct tokens across datasets, preventing shared representations. For example, a single drug (e.g., paracetamol) may map to one ATC code in US data but hundreds of billing codes in Taiwan claims. Without alignment, the model learns redundant embeddings, limiting generalization \citep{landi2020deep,finch2021exploiting}

We standardize all datasets into a unified coding space using ICD-10 (diagnoses), CPT (procedures), LOINC (labs), and ATC (drugs). Alignment is performed once at data build time, producing a fixed mapping dictionary shared across training and evaluation. We first construct an inventory of unmapped local codes by aggregating raw descriptions, English glosses, and heuristic matches. This yields a candidate pool of thousands of region-specific codes requiring alignment.

We then use an LLM such as Claude  to map each code to its closest standardized equivalent, returning a predicted code, confidence score, and short rationale. We retain only medium- and high-confidence mappings, and apply mappings in priority order: official mapping crosswalks, LLM-curated mappings, then TF-IDF fuzzy matching, with unmapped codes passed through unchanged. All aligned events are emitted as standardized tokens, ensuring consistent embeddings across datasets. More implementation details are in Appendix \ref{subsec:code_alignment}.

To evaluate few-shot cross-system generalization, we further align external EHR data from the EHRShot benchmark to our curated unified coding space. EHRShot contains heterogeneous coding systems, including SNOMED, RxNorm, CPT, and ICD variants, which differ substantially across healthcare settings \citep{wornow2023ehrshot}. Where available, we firstly leverage official mappings (e.g., CMS and NLM crosswalks) to map concepts into the unified space. For codes without explicit mappings, we formulate the mapping as a dense retrieval problem, a method widely used to align medical concepts across ontologies. We apply a textual semantic matching approach by embedding code descriptions and retrieving the most similar entries in the unified coding space with similarity threshold. This setting provides challenging benchmark for evaluating cross-system generalization of EHR foundation models. More implementation details provided in the Appendix \ref{subsec:code_alignment}.

\begin{figure}[t]
\centering

  \begin{minipage}[t]{0.48\textwidth}
      \captionof{table}{Statistics of unified dataset.}
         \vspace{-0.25cm}
    \label{tab:dataset_summary_statistics}
    \centering
     \resizebox{\textwidth}{!}{\begin{tabular}{l r} 
    \toprule
    \textbf{Item} & \textbf{Count} \\
    \midrule
    $\#$ of patients & 5,771,780 \\
    $\#$ of visits  & 1,493,521,343 \\
    $\#$ of diagnosis & 1,597,730,855 \\
    $\#$ of procedures  & 361,553,861 \\
    $\#$ of medications & 3,208,319,969 \\
    $\#$ of unique medical codes & 225,568 \\
    $\#$ of tokens for all codes & 9,479,408,953 \\
    $\#$ of patients for pretraining (80\%) & 4,617,424 \\
    $\#$ of patients for finetuning (20\%) & 1,654,356 \\
    Avg. $\#$ of visits per patient & 258.7 \\
    \bottomrule
    \end{tabular}}
  \end{minipage}%
 \hfill
\begin{minipage}[t]{0.48\textwidth}

\centering

\captionof{table}{Statistics of target cohort.}

\vspace{-0.2cm}

\label{tab:target_cohort}

\small

\resizebox{\textwidth}{!}{

\begin{tabular}{l l r r r}

\toprule

\textbf{Task} & \textbf{ICD Code} & \textbf{Positive} & \textbf{Total} & \textbf{Ratio (\%)} \\

\midrule

\multicolumn{5}{l}{\textit{Cancer}} \\

\midrule

Bladder  & C67.x & 3,946 & 1,535,483 & 0.26 \\

Breast   & C50.x & 17,153 & 1,548,690 & 1.11 \\

Colorectal & C18.x/C19/C20 & 11,092 & 1,542,629 & 0.72 \\

Kidney   & C64.x & 2,922 & 1,534,459 & 0.19 \\

Liver    & C22.x & 9,880 & 1,541,417 & 0.64 \\

Lung     & C34.x & 11,664 & 1,543,201 & 0.76 \\

\midrule

\multicolumn{5}{l}{\textit{Cardiovascular}} \\

\midrule

ACS & I25.x & 61,618 & 1,585,817 & 3.89 \\

CAD & I25.x & 153,466 & 1,569,778 & 9.78 \\

Cerebral Aneurysm & I67.1 & 18,693 & 1,591,896 & 1.17 \\

Heart Failure & I50.x & 109,185 & 1,586,097 & 6.88 \\

Stroke & I61.x/I63.x & 69,040 & 1,570,569 & 4.40 \\

\bottomrule

\end{tabular}}

\end{minipage}
\vspace{-0.6cm}
\end{figure}
\subsection{Training Data Construction}
\label{sec:data_curation}
\textbf{Overview.} We accessed a subset of National Health Insurance Research Database (NHIRD) through collaboration with a Taiwanese medical school, under IRB exemption. The dataset includes over 3 million patients from 1996–2013, covering 1.4B visits with diagnoses, procedures, and medications, providing more than 15 years of longitudinal, physician-validated records over 9B medical codes. Following code alignment, we integrate this data with 2 million US patients from multiple states to construct a unified dataset comprising over 9B standardized medical codes. All data were de-identified and not linked to any external information. Table \ref{tab:dataset_summary_statistics} summarizes the dataset characteristics.

\textbf{Target Cohorts.} For chronic disease prediction, we focus on major conditions including cancer: bladder, breast, colorectal, kidney, liver, and lung; and cardiovascular diseases: acute coronary syndrome (ACS), coronary artery disease (CAD), cerebral aneurysm, heart failure, and stroke. where early detection provides substantial clinical benefit~\citep{kukhareva2024lung, thiele2024screening, blackford2024pancreatic, haue2024artificial}. Each task is formulated as a binary classification problem, predicting whether a patient will develop the target disease. As suggested by clinicians, we predict disease onset one year after a given clinical visit to enable early intervention. We use up to five years of patient medical history as input, consistent with standard clinical practice. We define positive cohorts using ICD codes, taking the first occurrence of the target code. We randomly sample negative cohorts without disease-related codes and match them to positive patients by age and gender. To support broader disease coverage, we use top-level ICD codes to capture more cases. Statistics are shown in Table \ref{tab:target_cohort}.

\section{Methods}

This section presents the model, training, and inference of \method{}.

\subsection{Model}
\begin{figure*}[t]
    \centering
    \includegraphics[width=0.9\textwidth]{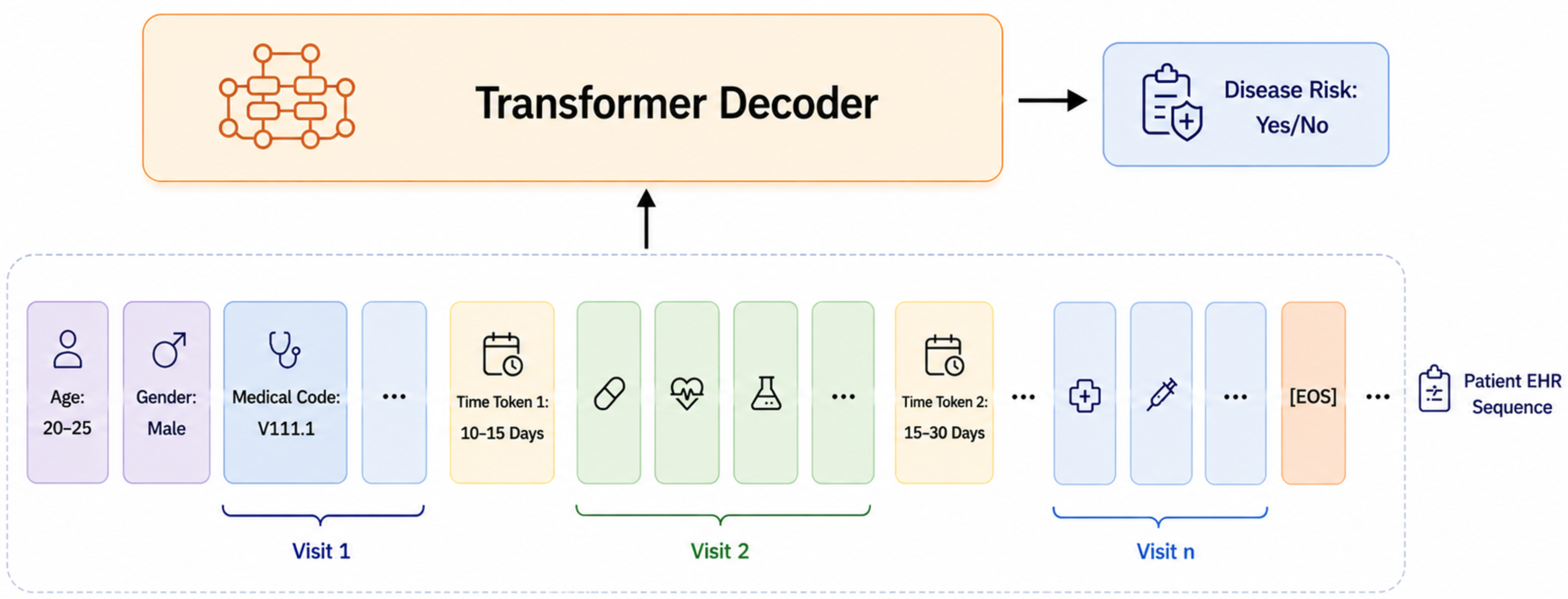}
    \vspace{-0.1cm}
    \caption{An example input sequence for a patient record and the architecture of \method.}
    \label{fig:architecture}
        \vspace{-0.5cm}
\end{figure*}

Figure~\ref{fig:architecture} illustrates the model setup of \method{}. 
It represents the healthcare record of each patient as a medical code sequence and models them by a decoder-only transformer.

\textbf{Patient Representation.} 
\label{CatchFM:medical-coding}
In longitudinal structured EHR databases, the patient records are often represented by a coding system, using a unique ID to encode each specific piece of information.
A typical patient record can start with their demographic information, such as age $c_{\text{age}}$ and gender $c_{\text{gender}}$, followed by a sequence of $n$ chronologically ordered visits,
${v_1, v_2, ..., v_n}$. Each visit $v_i$ consists of $m$ medical events, ${c_i^1, c_i^2, ..., c_i^m}$, covering diagnoses, medications, procedures, and other clinical events.

Specifically, \method{} represents each patient's EHR token sequence $\boldsymbol{x}$ as,
\begin{align}
[c_{\text{age}}, c_{\text{gender}}, &v_1, t_1, v_2, \ldots, v_{n-1},t_{n-1}, v_n, \texttt{[EOS]}]; & 
v_n &= [c_n^1, c_n^2, \ldots, c_n^m].
\end{align}

The demographic tokens age  \(c_{\text{age}}\) is discretized into predefined categorical ranges and gender \(c_{\text{gender}}\) is assigned as a distinct token. Each medical event code $c$ is assigned a unique token. 
Time intervals between consecutive visits are captured using time tokens $t$, which are also discretized into predefined categories to encode temporal information.  A special token, \(\texttt{[EOS]}\) marks the end of the record.

 Though each ID can be mapped into the language space, for example, to their names and descriptions, \method{} directly operates in the ID space which is more compact---each event is encoded by one token---and precise without ambiguity.

\textbf{Architecture.}  \method{} uses the standard decoder-only transformer architecture $G_{\boldsymbol{\theta}}$ on top of the patient's EHR token sequence $\boldsymbol{x}$. 
We use rotary positional embeddings (RoPE)~\citep{su2023roformerenhancedtransformerrotary} to encode positional information for visits:
\begin{align}
\text{RoPE}(\boldsymbol{x_i}, p_i) = \boldsymbol{h_i} \cdot \cos(\theta( p_i)) + \boldsymbol{h_{i\perp}} \cdot \sin(\theta( p_i)).    
\end{align}
It assigns the same absolute position $p_i$ (i.e., replacing sequential positions like 0,1,2,3... with 0,0,1,1...) to event tokens $\boldsymbol{x_i}$ from the same visit. This allows the model to capture the relationships between events within and across visits.

\subsection{Pretraining, Finetuning, and Inference}

\method{} employs the standard pretraining, finetuning, and inference pipeline.

\textbf{Pretraining.}
\label{subsec:pretrain}
 We pretrain  \method{} $G_{\boldsymbol{\theta}}$ \textit{from scratch} on our healthcare record pretraining dataset, using the standard autoregressive next-token prediction objective:
\begin{align}
\min_{\boldsymbol{\theta}}\mathcal{L}_{\text{LM}},~~ \mathcal{L}_{\text{LM}} &= -\frac{1}{n}\sum_{j=1}^n \log f_{\boldsymbol{\theta}}(x_j|\boldsymbol{x}_{<j}), & 
    f_{\boldsymbol{\theta}}(x_j|\boldsymbol{x}_{<j}) &= \text{Softmax} (\boldsymbol{E}  \boldsymbol{h}_j).
\end{align}
The next code probability $f_{\boldsymbol{\theta}}(\cdot)$ is computed on token embeddings $\boldsymbol{E}$ and hidden states $\boldsymbol{h}$ from $G_{\boldsymbol{\theta}}$. 
Pretraining on large scale patient record with autoregressive language modeling task enables  \method~to capture the complex medical patterns in patient health trajectories, for example, associations of medical events and potential risk factors of diseases.

\textbf{Finetuning.} 
\label{subsec:sft} For chronic disease prediction task, we employ
supervised fine-tuning on the pretrained \method~$G_{\boldsymbol{\theta}}$.  It uses cross-entropy loss to learn whether a patient will be diagnosed as a target disease outcome $y$.
\begin{align}
    \min_{\boldsymbol{\theta},\boldsymbol{\phi}}\mathcal{L}_{\text{SFT}},~~ \mathcal{L}_{\text{SFT}} &= - \log f_{\boldsymbol{\phi}} (y \mid \boldsymbol{x}_{\texttt{[EOS]}}), &
    f_{\boldsymbol{\phi}}(y \mid \boldsymbol{x}_{\texttt{[EOS]}}) &= \text{Softmax} (\boldsymbol{W} \boldsymbol{h}_{\texttt{[EOS]}}+\boldsymbol{b}).
\end{align}
The learnable parameters include $\theta$ in the foundation model, and $\boldsymbol{W}$ and $\boldsymbol{b}$ from a linear prediction layer. Finetuning makes \method{} specialized for disease risk prediction by capturing risk factors from medical histories while leveraging general medical knowledge learned from pretraining.

\textbf{Inference.} The inference of \method{} is to take a patient's medical history from their EHR record, run a forward pass of the pretrained and finetuned \method{} instances, and predict the disease risk of the patient. The inference is efficient as only one prediction token is needed per patient. 

The reliance on patient history alone makes \method{} a natural fit for chronic disease prediction in population health management. Healthcare providers can deploy \method{} at scale on large EHR datasets to identify high-risk patients. These patients can then be further evaluated by clinicians to determine appropriate timing and need for intervention. Accurate risk prediction enables effective triage, improving the efficiency of care delivery and providing a larger window for timely treatment and better patient outcomes.

\begin{figure*}[t]
    \centering
    \begin{minipage}[t]{0.7\textwidth}
    \begin{subfigure}[t]{0.33\textwidth}
        \centering
        \includegraphics[width=\textwidth]{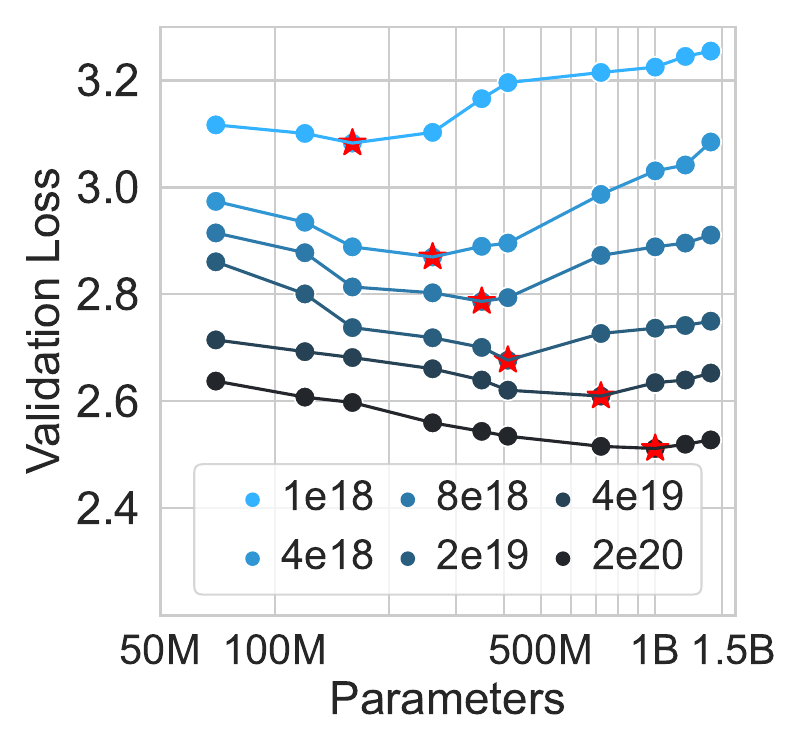}
        \caption{Loss Scaling}
        \label{fig:scale_a}        
    \end{subfigure}
    \begin{subfigure}[t]{0.33\textwidth}
        \centering
    \includegraphics[width=\textwidth]{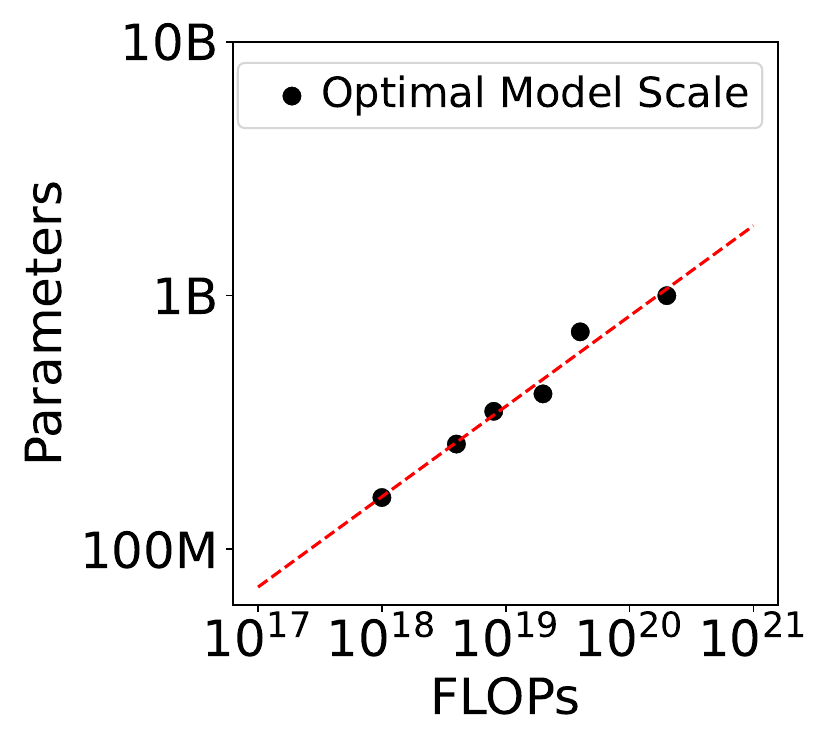}
        \caption{Model Scaling}
        \label{fig:scale_b}
    \end{subfigure}%
    \begin{subfigure}[t]{0.33\textwidth}
        \centering
    \includegraphics[width=\textwidth]{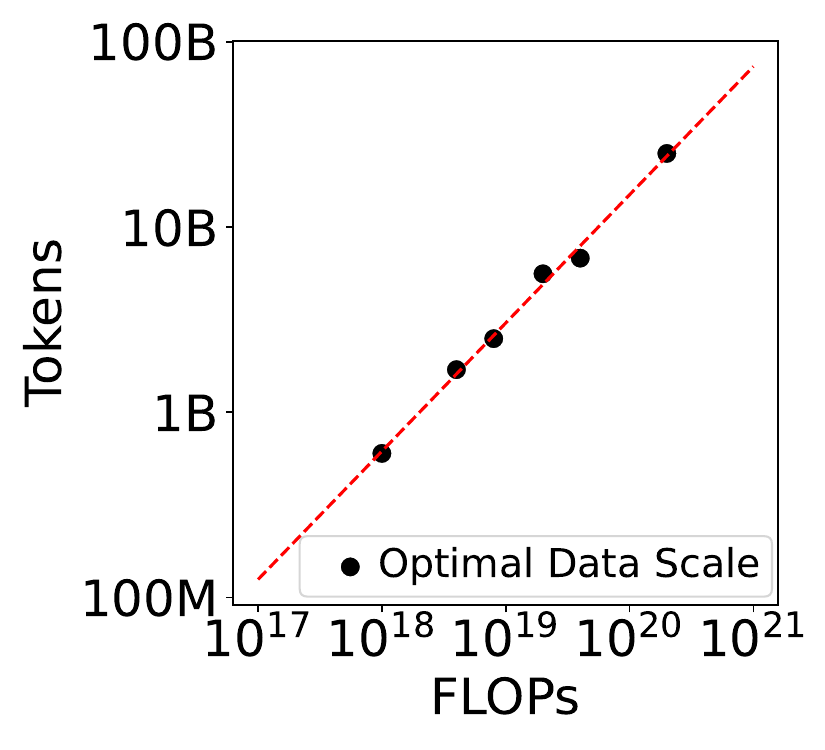}
        \caption{Token Scaling}
        \label{fig:scale_c}
    \end{subfigure}%
\caption{Scaling Law of \method{}, including loss IsoFLOPs,  estimated FLOP-optimal parameters and pretraining tokens.}
	\label{fig:scaling_law}
    \end{minipage}
    \hfill
    \begin{minipage}[t]{0.28\textwidth}
    \vspace{-2.9cm}
    \centering
    \captionof{table}{Compute-optimal \method{} instances pretrained with variant FLOPs, and their estimated GPU hours in typical A100 machines.}
    \label{tab:gpu_hours}
    \resizebox{\textwidth}{!}{%
        \begin{tabular}{rrr}
        \toprule
    \textbf{FLOPs}&\textbf{Parameters} & \textbf{A100 Hours} \\
        \midrule
        1e18  & 160m&$\sim$16 \\
        4e18   & 260m&$\sim$64 \\
        2e19  & 410m&$\sim$128 \\
        4e19  & 720m&$\sim$256 \\
        2e20 & 1b&$\sim$1280 \\
        \bottomrule
        \end{tabular}
    }        
    \end{minipage}
    \vspace{-0.3cm}
\end{figure*}
\section{Experimental Methodologies}
\label{sec:experiment-setup}
 \textbf{Dataset.} We use the curated dataset described in Sec.~\ref{sec:data_curation} for pretraining, finetuning, and cross-system evaluation. The dataset is randomly partitioned into 80\% for pretraining and 20\% for downstream chronic disease prediction task, with no patient overlap. Each task-specific dataset is further split into 80\%/10\%/10\% for train/validation/test using stratified sampling to preserve class balance across target disease cohorts.

In addition, we evaluate few-shot generalization on the public EHRSHOT benchmark~\citep{wornow2023ehrshot}, which contains 6,739 Stanford Medicine patients (1990–2023) coded primarily in SNOMED, CPT, and RxNorm. After code alignment, we evaluate \method{} on six new diagnosis prediction tasks: acute myocardial infarction, lupus, hyperlipidemia, hypertension, celiac disease, and pancreatic cancer, following the standard few-shot setting (1–128 positive examples).


\textbf{Evaluation metrics.}  Task performance is evaluated using AUROC, AUPRC, specificity, and sensitivity, with AUROC and AUPRC as the primary metrics. AUROC measures the model’s ability to discriminate between patients with and without disease across all thresholds, while AUPRC captures the precision–recall trade-off under class imbalance. Specificity (true negative rate) reflects reliability in identifying low-risk patients, and sensitivity (true positive rate) measures the model’s ability to detect high-risk patients. When applicable, we report sensitivity at 99\% specificity, a clinically relevant operating point that reduces false positives and improves reliability~\citep{cao2023large, Halner2023_DEcancer, Jopek2025_ImpactOC}.

\textbf{Baselines.} We include standard tree-based models with bag-of-words inputs~\citep{bharati2023review}, including XGBoost~\citep{chen2016xgboost} and LightGBM~\citep{ke2017lightgbm}. We also compare against pretrained language models, BioMedLM~\citep{bolton2024biomedlm27bparameterlanguage} and Qwen2.5~\citep{yang2024qwen2}, by converting medical codes into text and finetuning on the same data. For EHRSHOT, we benchmark \method{} against methods reported on the official leaderboard~\citep{wornow2023ehrshot}.

\textbf{Implementation Details.} We implement \method{} using a standard architecture, Pythia~\citep{biderman2023pythiasuiteanalyzinglarge}, a decoder-only transformer with rotary position embeddings and flash attention, and pretrain from scratch on unified data. Medical codes, demographics, and special tokens form a 225,568 token vocabulary, with sequences capped at 2,048 tokens. Longer histories are chunked without overlap during pretraining and truncated from left side for fine-tuning. More architecture, hyperparameters, and training details are provided in Appendix~\ref{sec:implement}.

\section{Evaluation Results}

This section presents experimental results evaluating \method{}'s scaling law, effectiveness, cross-site generalization, and ablation studies.

\subsection{Scaling Laws of Pretraining on Healthcare Records}
\label{sec:scaling_laws}

To pretrain effectively on healthcare records, we first conduct a thorough scaling law analysis of \method{}.
Specifically, we use IsoFLOP profiling~\citep{hoffmann2022training}: pretraining models with various sizes at target FLOPs by varying the number of pretraining tokens.

Figure~\ref{fig:scale_a} illustrates clear parabola-shaped IsoFLOP curves, with different model sizes achieving minimum validation loss at various FLOPs. 
Using these data points, we fit a power law to characterize the relationship between FLOPs $(C)$, the loss-optimal model size $(N_\text{opt})$, and the optimal number of training tokens $(D_\text{opt})$, as illustrated in Figures \ref{fig:scale_b} and \ref{fig:scale_c}:
\begin{align}
   &\text{Optimal Model Sizes: }  N_{\text{opt}} \propto C^{0.34},
   & {  } &\text{Optimal Token Counts: }  D_{\text{opt}} \propto C^{0.69}.
\end{align}

Scaling laws in healthcare foundation models resemble the scaling laws of large language models~\citep{hoffmann2022training}, highlighting the potential of large-scale EHR  foundation models. While emergent capabilities need further study, we pretrain and finetune a series of compute-optimal \method{} models across various FLOPs (Table~\ref{tab:gpu_hours}) and evaluate them in chronic disease prediction. 

\begin{table*}[t]
\centering
\caption{Performance comparison on the US and NHIRD datasets. Each entry is reported as AUROC/AUPRC/sensitivity at 99\% specificity. The best value for each row is highlighted.}
\label{tab:overall}
\resizebox{\textwidth}{!}{
\begin{tabular}{lrrrrrrr}
\toprule
\textbf{Task} & \textbf{XGBoost} & \textbf{LightGBM} & \textbf{BioMedLM-2.7b} & \textbf{Qwen2.5-3b} & \textbf{EHR-FM 160m} & \textbf{EHR-FM 1b} & \textbf{EHR-FM 2.4b} \\
\midrule

\multicolumn{8}{l}{\textbf{Population:} US} \\
\midrule
Cancer Bladder & 81.9/2.9/12.5 & 81.3/1.4/10.7 & 59.7/1.4/10.1 & 86.6/2.5/23.8 & 88.2/3.6/27.4 & 89.6/4.3/31.5 & \textbf{91.1/7.5/29.8} \\
Cancer Breast & 89.6/11.6/20.1 & 89.9/11.4/19.9 & 76.3/6.1/16.1 & 89.5/12.0/26.9 & 91.8/17.7/32.6 & 93.0/20.3/34.9 & \textbf{93.3/22.1/37.6} \\
Cancer Colorectal & 75.9/1.6/8.3 & 78.7/1.2/7.6 & 76.4/1.8/12.9 & 80.3/2.7/17.2 & 82.8/4.3/16.2 & 85.4/5.4/20.1 & \textbf{86.5/5.0/20.8} \\
Cancer Kidney & 69.7/1.3/14.5 & 80.2/0.9/6.2 & 56.0/0.4/5.6 & 81.7/1.7/10.6 & 84.4/3.5/21.8 & 86.0/4.5/25.7 & \textbf{87.5/4.3/21.2} \\
Cancer Liver & 65.6/2.6/32.0 & 74.3/3.1/16.5 & 69.2/0.5/8.7 & 86.5/2.5/27.2 & 91.8/4.8/39.8 & 92.4/8.7/43.7 & \textbf{93.1/8.9/42.7} \\
Cancer Lung & 70.5/2.2/7.4 & 87.3/4.3/16.4 & 79.5/1.9/10.1 & 85.9/3.4/14.6 & 89.6/6.2/19.7 & 90.9/9.0/26.9 & \textbf{91.4/9.9/28.7} \\
Acute Coronary Syndrome & 89.9/50.8/39.8 & 91.5/54.9/43.6 & 86.5/35.8/25.7 & 89.0/44.8/33.4 & 93.6/65.3/54.6 & 94.5/68.2/58.1 & \textbf{95.0/70.0/59.0} \\
Coronary Artery Disease & 89.4/57.2/30.7 & 91.2/62.6/35.7 & 89.1/54.0/26.2 & 90.4/58.6/30.3 & 93.7/71.6/45.8 & 94.6/75.1/49.6 & \textbf{95.0/76.1/51.2} \\
Cerebral Aneurysm & 85.0/25.9/33.0 & 86.7/26.9/36.4 & 59.5/1.3/3.6 & 81.2/17.8/26.5 & 88.1/35.6/44.7 & 91.1/41.9/49.9 & \textbf{91.8/43.0/50.9} \\
Heart Failure & 91.5/60.4/39.1 & 92.9/64.9/43.2 & 89.8/51.4/29.1 & 91.7/58.5/35.7 & 94.8/73.9/55.0 & 95.5/76.8/58.2 & \textbf{95.9/78.1/60.3} \\
Stroke & 87.6/42.0/33.5 & 89.5/46.7/37.0 & 73.9/19.4/16.0 & 87.6/39.3/30.2 & 92.1/57.7/48.1 & 93.4/62.2/53.0 & \textbf{93.7/63.6/54.0} \\
\textbf{Average} & 81.5/23.5/26.9 & 85.8/25.3/24.9 & 74.2/15.8/15.9 & 86.4/22.2/25.1 & 90.1/31.3/36.9 & 91.5/34.2/41.0 & \textbf{92.2/35.3/41.5} \\

\midrule

\multicolumn{8}{l}{\textbf{Population:} NHIRD}  \\
\midrule
Cancer Bladder & 91.2/16.1/32.6 & 94.5/33.9/56.1 & 82.7/2.9/13.0 & 85.7/3.8/17.4 & 92.6/54.7/62.2 & 91.5/56.6/59.1 & \textbf{92.3/58.8/62.6} \\
Cancer Breast & 93.4/27.2/31.9 & 95.3/33.5/40.3 & 91.4/26.3/31.1 & 93.3/29.8/38.0 & 95.7/59.7/61.4 & 93.9/64.9/64.4 & \textbf{94.9/65.1/64.5} \\
Cancer Colorectal & 88.4/16.2/20.3 & 91.0/24.1/29.7 & 84.1/11.1/15.6 & 87.2/13.1/16.4 & 93.4/57.4/57.5 & 92.7/63.1/63.1 & \textbf{90.7/63.6/61.8} \\
Cancer Kidney & 91.5/23.9/38.5 & 94.4/45.7/69.7 & 75.1/4.7/13.1 & 87.2/5.6/27.1 & 92.5/63.4/68.9 & 94.1/66.1/68.9 & \textbf{93.2/67.4/71.3} \\
Cancer Liver & 91.8/38.0/41.0 & 94.1/42.7/47.7 & 89.2/29.0/32.8 & 90.2/29.3/30.7 & 95.5/69.0/70.4 & 95.1/73.7/72.1 & \textbf{95.2/74.7/73.2} \\
Cancer Lung & 86.9/17.6/7.4 & 90.8/26.3/35.3 & 81.1/7.8/12.6 & 84.0/10.2/14.9 & 91.6/57.8/59.6 & 89.3/63.3/63.2 & \textbf{90.9/62.5/60.0} \\
Acute Coronary Syndrome & 92.4/56.8/45.1 & 94.2/62.3/50.3 & 89.2/41.4/29.5 & 92.4/52.7/39.5 & 97.4/85.7/81.0 & 97.4/87.6/84.6 & \textbf{97.7/87.7/83.4} \\
Coronary Artery Disease & 93.1/73.1/45.7 & 94.7/77.9/52.1 & 90.5/66.4/38.1 & 93.5/73.6/46.4 & 97.9/91.6/81.5 & 98.0/92.8/85.6 & \textbf{98.2/93.3/86.3} \\
Cerebral Aneurysm & 88.1/29.3/30.9 & 91.1/35.7/37.0 & 71.4/7.4/9.0 & 87.0/24.3/24.5 & 95.1/74.3/73.4 & 96.3/78.8/76.8 & \textbf{95.7/79.7/78.1} \\
Heart Failure & 92.9/62.4/41.2 & 94.3/67.9/47.9 & 87.3/40.8/21.1 & 93.0/60.9/37.7 & 97.8/87.9/79.1 & 97.8/89.3/81.8 & \textbf{98.2/90.5/84.3} \\
Stroke & 91.7/58.8/40.4 & 93.5/64.7/46.1 & 70.5/15.2/6.7 & 92.8/61.1/42.0 & 96.7/85.0/77.0 & 97.0/87.5/81.5 & \textbf{97.6/89.2/83.5} \\
\textbf{Average} & 91.1/38.1/34.0 & 93.4/46.8/46.3 & 83.0/23.0/20.2 & 89.6/34.9/30.7 & 95.1/71.5/70.2 & 95.0/74.9/72.8 & \textbf{95.0/75.7/73.6} \\

\bottomrule
\end{tabular}
}
\end{table*}
\subsection{Overall Results} 
\label{sec:overall_results}

Table~\ref{tab:overall} shows the overall performance of chronic disease prediction in NHIRD and US dataset.
\method{} pretrained on aligned cross-site dataset outperforms XGBoost and LightGBM, strong tree models, by 10\%+ AUPRC on US target disease cohorts and 30\%+ AUPRC on NHIRD target disease cohorts.
It also significantly outperforms medical (BioMedLM) and general (Qwen2.5) large language models finetuned on the texts of patient EHR inputs.  Sect.~\ref{sec:ablation} further studies the benefits of pretraining directly on medical codes rather than converting codes into natural language. 
Moreover, our studies show that \method{} achieve 40\% and 70\% sensitivity in US and NHIRD cohorts at a decision threshold corresponding to 99\% specificity,  confirming its ability to identify high-risk patients for further screening (high sensitivity) and avoid unnecessary patient distress (high specificity) for population health management.


Chronic disease prediction is a challenging task, particularly for cancer prediction. \method{} demonstrates strong effectiveness, achieving 30\%+ and 65\%+  sensitivity at 99\% specificity in predicting cancer on US and NHIRD cohorts. The benefit of scale is evident in all cohorts. \method-2.4b outperforms \method-160m by 5\% sensitivity on US and NHIRD cohorts. Scaled-up foundation models perform better on the harder cancer prediction task.

\begin{figure*}[t]
\centering
\vspace{-0.2cm}

\begin{subfigure}[t]{0.32\linewidth}
    \centering
    \includegraphics[width=\linewidth]{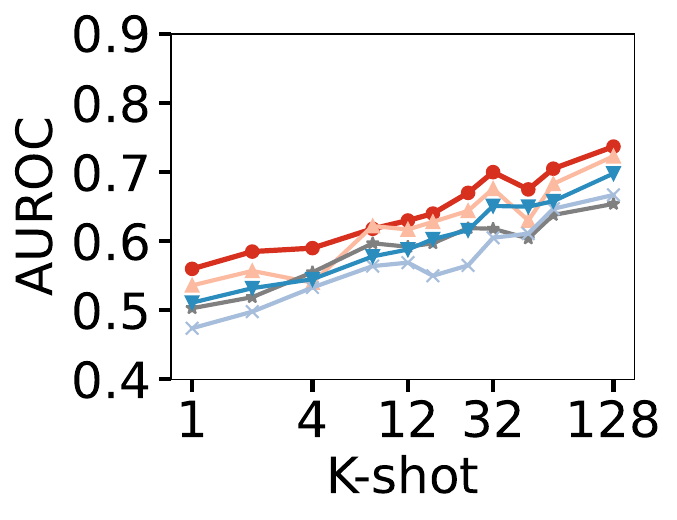}
    \caption{Hyperlipidemia}
\end{subfigure}
\hfill
\begin{subfigure}[t]{0.32\linewidth}
    \centering
    \includegraphics[width=\linewidth]{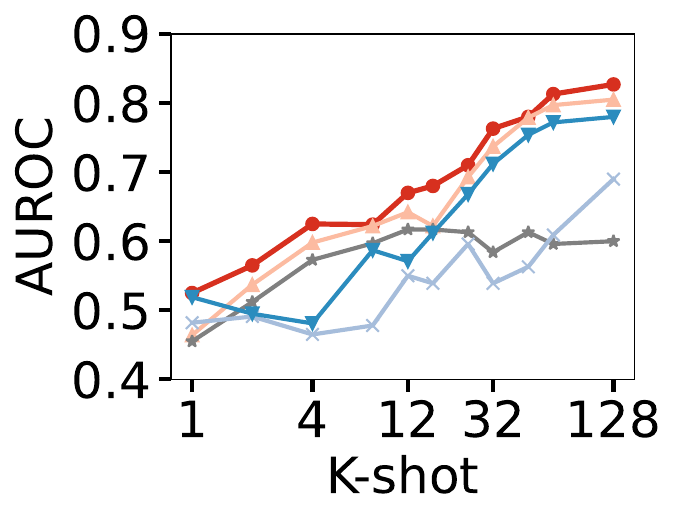}
    \caption{Celiac}
\end{subfigure}
\hfill
\begin{subfigure}[t]{0.32\linewidth}
    \centering
    \includegraphics[width=\linewidth]{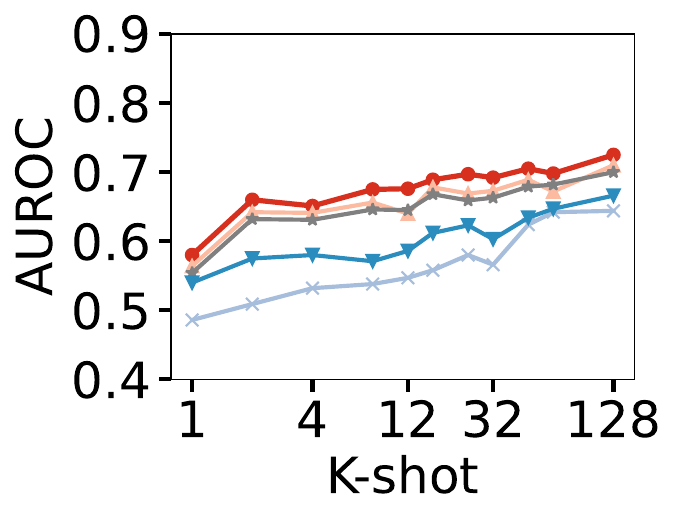}
    \caption{Hypertension}
\end{subfigure}

\vspace{0.2cm}

\begin{subfigure}[t]{0.32\linewidth}
    \centering
    \includegraphics[width=\linewidth]{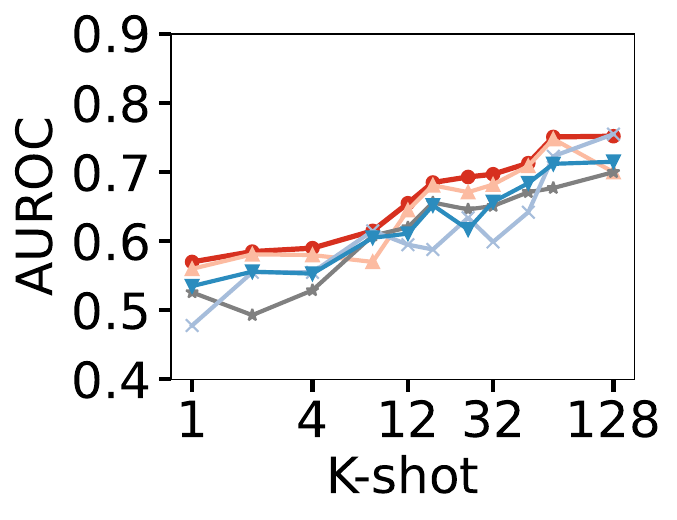}
    \caption{Lupus}
\end{subfigure}
\hfill
\begin{subfigure}[t]{0.32\linewidth}
    \centering
    \includegraphics[width=\linewidth]{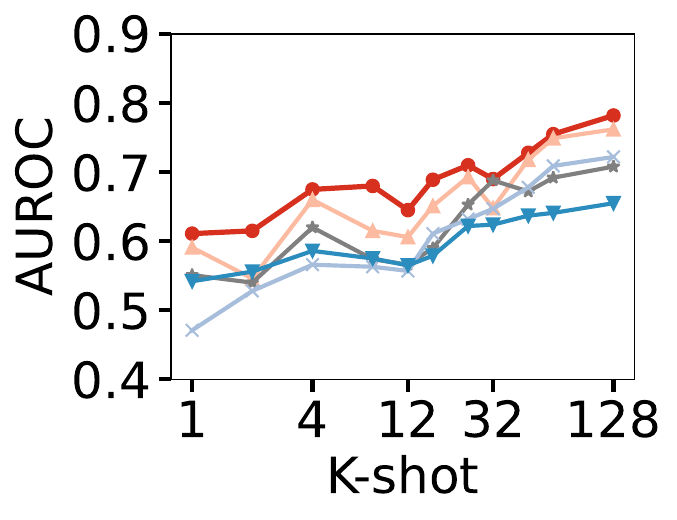}
    \caption{Acute Myocardial Infarction}
\end{subfigure}
\hfill
\begin{subfigure}[t]{0.32\linewidth}
    \centering
    \includegraphics[width=\linewidth]{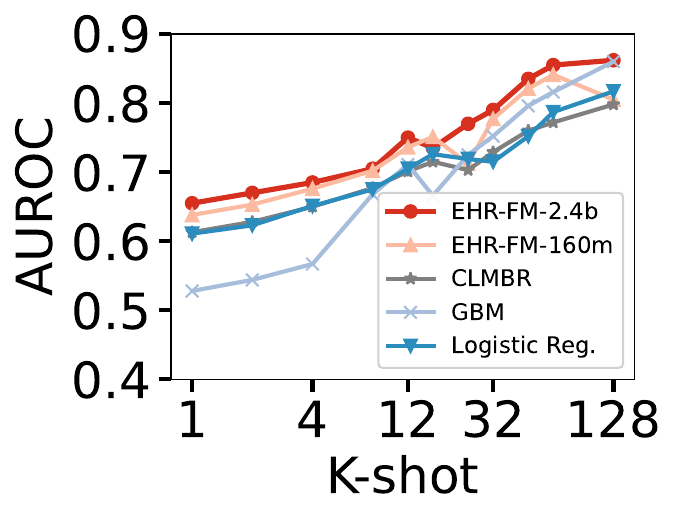}
    \caption{Pancreatic Cancer}
\end{subfigure}

\caption{AUROC on the EHRSHOT benchmark for new diagnosis assignment tasks.}
\label{fig:ehrshot_auroc}
\vspace{-0.3cm}
\end{figure*}
\begin{table*}[t]
\centering
\caption{Performance comparison across different mixtures of pretraining data on the US cohort.}
\label{tab:data_limited}
\small
\resizebox{\textwidth}{!}{\begin{tabular}{lrrrr}
\toprule
\textbf{Task} & \textbf{No pretrained} & \textbf{NHIRD} & \textbf{US} & \textbf{ US + NHIRD} \\
\midrule
Cancer Bladder              & 85.2/3.3/23.2  & 88.9/3.6/27.4  & 89.9/5.2/31.0  & \textbf{90.5/7.0/34.5} \\
Cancer Breast               & 91.2/15.5/30.5 & 92.3/19.3/33.5 & 92.5/19.0/33.8 & \textbf{92.9/20.9/36.2} \\
Cancer Colorectal           & 82.4/2.7/18.5  & 83.7/4.7/18.2  & 84.4/4.0/20.1  & \textbf{85.8/5.0/19.5} \\
Cancer Kidney               & 81.1/1.6/19.0  & 81.8/1.7/17.3  & \textbf{86.1/5.5/24.6}  & 83.8/2.6/20.7 \\
Cancer Liver                & 87.9/6.6/39.8  & 88.9/5.5/33.0  & 91.1/4.9/40.8  & \textbf{91.4/5.6/43.7} \\
Cancer Lung                 & 88.1/6.0/20.9  & 89.5/7.4/24.4  & 90.6/7.4/24.0  & \textbf{90.8/8.0/24.2} \\
Acute Coronary Syndrome     & 92.2/60.4/50.6 & 93.6/65.5/55.2 & 94.1/65.6/54.7 & \textbf{94.2/67.5/56.9} \\
Coronary Artery Disease     & 93.0/69.6/44.0 & 93.7/71.9/46.1 & 93.9/72.9/46.8 & \textbf{94.3/73.9/48.2} \\
Cerebral Aneurysm           & 86.7/28.5/35.7 & 88.9/35.4/44.1 & 90.4/38.2/47.9 & \textbf{90.7/39.2/49.5} \\
Heart Failure               & 94.2/71.6/51.4 & 94.9/74.3/54.8 & 95.1/74.7/55.5 & \textbf{95.4/76.3/57.3} \\
Stroke                      & 90.7/53.3/44.0 & 91.9/58.2/49.1 & 92.7/59.7/50.0 & \textbf{93.0/61.2/51.9} \\
\textbf{Average}                & 88.4/29.0/34.3 & 89.8/31.6/36.6 & 91.0/32.5/39.0 & \textbf{91.2/33.4/40.2} \\
\bottomrule
\end{tabular}}
\end{table*}
\subsection{Generalization Ability}
We evaluate \method{} under two generalization settings: few-shot adaptation and data-limited pretraining.
\label{sec:generalization}

\textbf{Few-shot Adaptation.} Figure~\ref{fig:ehrshot_auroc} shows the K-shot performance of \method{} on the EHRSHOT for new diagnosis prediction, using K positive and K negative on-site examples for training. Despite substantial distribution shifts in population and healthcare systems, and over 20\% medical code mismatch prior to alignment, \method{} achieves state-of-the-art performance after code alignment with only a handful of on-site examples, maintaining the best under few-shot generalization.

Moreover, \method-160m consistently outperforms CLMBR (140M) in AUROC across nearly all K-shot. Scaling to \method-2.4b further improves few-shot performance. Notably, CLMBR is pretrained at a similar scale using on-site data (2.57M Stanford patients, same as EHRSHOT), whereas \method{} operates under significant distribution shift, with only partial code overlap prior to alignment. Overall, these results show \method{}’s robustness to distribution shifts and code mismatch, underscoring its potential for transfer across healthcare systems. 


\textbf{Data-limited Pretraining.} To evaluate the effectiveness of unified cross-system pretraining, we consider three settings under the same token budget on US downstream cohorts with EHR-FM 160m: (1) 6B unique NHIRD-only tokens; (2) 100M unique US-only tokens repeated 60 times; and (3) a mixture of 100M unique US tokens (repeated 30 times) and 3B unique NHIRD tokens.

As shown in Table~\ref{tab:data_limited}, pretraining on NHIRD alone already outperforms the no-pretraining baseline on US downstream tasks, despite the distribution mismatch, demonstrating the benefit of curated pretraining after code alignment. Comparing settings (2) and (3), we observe that incorporating aligned cross-system data from NHIRD can further improve downstream performance beyond duplicating single-site US data alone, particularly under limited US pretraining data. These results suggest that aligned cross-system data provides an effective pretraining signal at scale, and is critical for further scaling EHR dataset and improving chronic disease prediction in population health settings.

\subsection{Ablation Study}
\label{sec:ablation}
\begin{figure}[t]
  \centering
\begin{minipage}[t]{0.47\textwidth}
    \centering
     \vspace{-3.2cm}
    \captionof{table}{Ablation study on various pretraining strategies and model architecture on cancer breast task at 160m with US cohorts.}
    \label{tab:ablation}
    \resizebox{\textwidth}{!}{%
\begin{tabular}{lrrr}
\toprule
\textbf{Variations} & \textbf{AUROC} & \textbf{AUPRC} & \textbf{Sensitivity} \\
\midrule
\multicolumn{4}{l}{\textbf{Medical Code Representation}} \\
\method & \textbf{91.8} & \textbf{17.7} & \textbf{32.6} \\
demographic-only features & 83.7 & 2.8 & 0.0 \\
No Pretrain & 90.1 & 15.3 & 28.7 \\
\multicolumn{4}{l}{\textbf{Model Architecture}} \\
\text{ } w/. Token-level Relative Position & 91.6 & 17.5 & 32.4 \\
\text{ } w/o. Time Tokens                 & 90.3 & 16.7 & 30.9 \\
\midrule
\multicolumn{4}{l}{\textbf{Converted Text Representation}} \\
demographic-only features & 82.7 & 2.1 & 0.0 \\
No Pretrain & 89.9 & 13.5 & 27.6 \\
Finetune Pythia & 89.8& 13.4& 28.0 \\
Pretrain from Scratch & 88.7 & 11.4 & 24.6 \\
Continual Pretrain Pythia & 91.0 & 15.1 & 27.7 \\
\bottomrule
\end{tabular}
    }
  \end{minipage}
    \hfill
  \begin{minipage}[t]{0.47\textwidth}
    \centering
    \setcounter{subfigure}{0}
    \begin{subfigure}[t]{0.47\textwidth}
      \includegraphics[width=\textwidth]{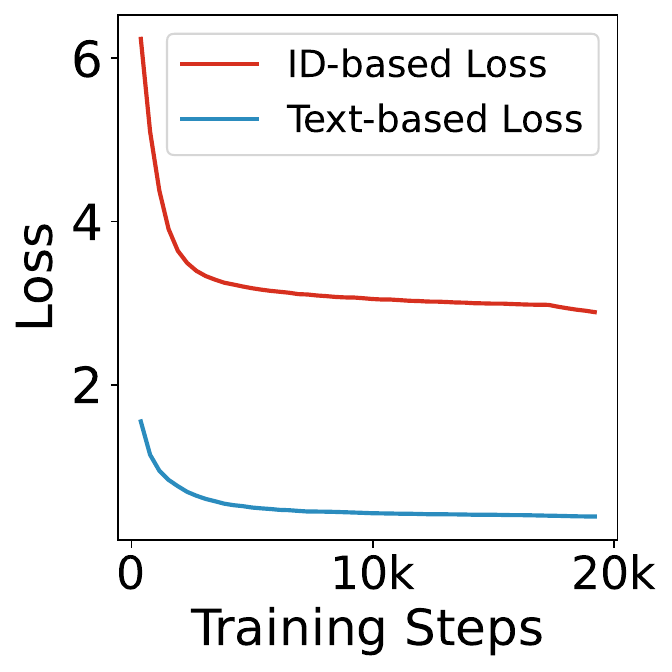}
      \vspace{-0.4cm}
      \caption{Overall}\label{fig:combined_loss}
    \end{subfigure}%
    \hfill
    \begin{subfigure}[t]{0.48\textwidth}
      \includegraphics[width=\textwidth]{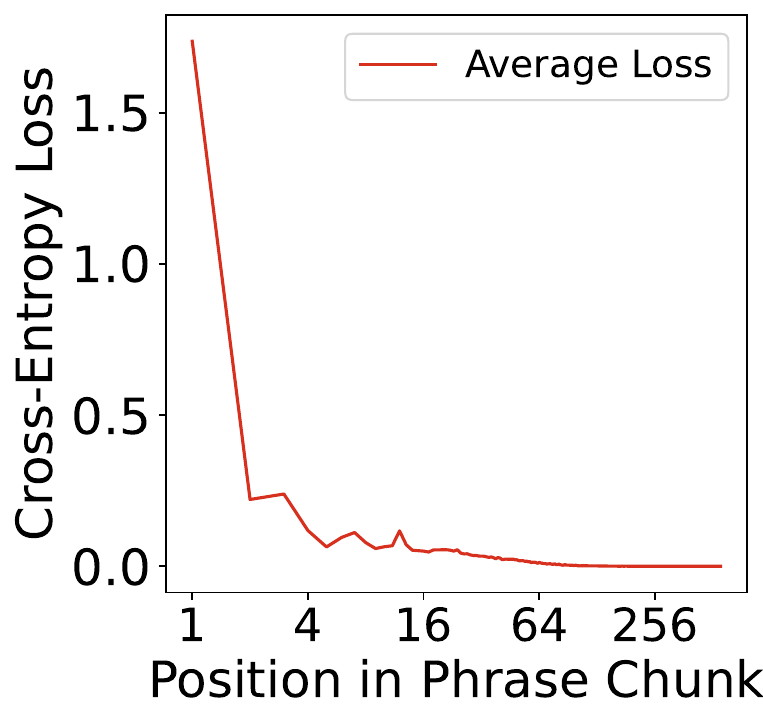}
      \vspace{-0.4cm}
      \caption{At Positions}\label{fig:chunk_loss}
    \end{subfigure}
    \caption{Pretraining losses of \method{} (a) on medical code IDs and texts and (b) at relative text token position in ID name phrase chunk.}\label{fig:loss}
  \end{minipage}
\end{figure}

Table~\ref{tab:ablation} shows ablation studies on different components of \method{}. Modeling patient history is a key source of evidence. Demographic information alone is a poor indicator; pretraining is a key advantage of \method{}; different model architectures, similar to observations in large language models, yield mild differences.

Converting the medical codes into their corresponding textual names and pretraining \method{} on texts hinders performance. As shown in Figure~\ref{fig:combined_loss}, the loss quickly drops to near zero, a strong sign of overfitting. The model memorizes code names and predicts them trivially after seeing the initial tokens. Figure~\ref{fig:chunk_loss} shows the loss is only meaningful on the first token, with the rest memorized. How to better adapt general domain LLMs to EHR data is an interesting future research direction.



\section{Conclusion}
\method{} provides a scalable approach for chronic disease risk prediction by pretraining and adapting Transformer-based models on large-scale longitudinal medical code sequences. Its strong predictive performance (over 40\% and 70\% sensitivity at 99\% specificity across U.S. and Taiwan cohorts), low false-positive rates, and applicability using routinely collected EHR data make it well-suited for population health management. By enabling early identification of high-risk individuals, \method{} can support clinicians in prioritizing screening and intervention, improving the efficiency and coverage of preventive care. Our experiments demonstrate the benefits of scaling EHR foundation models, their robustness to distribution shifts across heterogeneous healthcare systems, and the importance of unified code alignment for effective cross-system generalization. We further show that aligned cross-system data provides a stronger pretraining signal than duplicating single-site data, especially in data-limited medical settings. We hope our findings and open-source resources will facilitate future research and real-world deployment of EHR foundation models for large-scale, proactive healthcare.

\bibliographystyle{abbrvnat}
\bibliography{reference}
\newpage

\appendix
\section*{Impact Statement}
\label{sec:impact_statement}
\method{} could help improve population health management for chronic diseases by enabling large-scale risk stratification from longitudinal EHR data. Using routinely collected diagnoses, procedures, medications, and laboratory records, the model can identify patients who may be at elevated risk for future disease onset or progression. Such approaches could support earlier monitoring and preventive interventions for chronic conditions including cardiovascular diseases and cancer, where timely care is associated with improved outcomes.

In addition, \method{} could help healthcare systems prioritize limited clinical resources by identifying patients who could benefit from additional follow-up screening or care management. Because the framework is trained on longitudinal data across multiple healthcare systems, it may also improve the robustness and generalizability of EHR-based prediction models across institutions. While further prospective validation is necessary before clinical deployment, these results suggest that scalable EHR foundation models could become useful tools for supporting preventive and population-level healthcare strategies.

When integrated into real-world clinical workflows, \method{} could support value-based healthcare by reducing preventable hospitalizations, complications, and late-stage disease diagnoses. Combined with downstream clinical decision support tools and multimodal diagnostics such as medical imaging, it has the potential to improve healthcare accessibility, optimize medical resource utilization, and enable earlier, more effective interventions for chronic disease management at scale.

\section*{Limitation}
\label{sec:limitation}
Due to computational constraints, our study focuses on models up to 2.4 billion parameters. Although these models already achieve strong performance, it remains unclear whether scaling trends observed in general-domain language models continue to hold for large-scale healthcare foundation models at substantially larger scales such as Chinchilla (70B) \citep{hoffmann2022training} and GPT-3 (175B) \citep{brown2020languagemodelsfewshotlearners}. Future work should explore larger architectures and investigate how scaling affects clinical generalization, robustness, and efficiency across diverse healthcare systems.

In addition, our current framework relies on a unified coding space constructed through a combination of official mappings and text-based semantic alignment methods. While this approach substantially improves cross-site generalization, medical code alignment remains an open challenge due to differences in coding standards, healthcare practices, and local billing systems across institutions and countries. Future work should investigate alternative alignment strategies, including ontology-aware methods, learned alignment approaches, and multimodal representations incorporating clinical text or laboratory measurements, to further improve cross-system generalization and representation quality. It is also worth investigating methods for combining structured codes with free-text notes and possibly other data
types (e.g., lab results, imaging reports) within a unified pretraining framework.

\section{Data Curation}
\begin{table}[h]
\centering
\small
\caption{Code consolidation statistics after cross-system code alignment.}
\label{tab:code_alignment}
\begin{tabular}{lccc}
\toprule
\textbf{Mapping} & \textbf{Taiwan-side Codes} & \textbf{International Codes} & \textbf{Compression} \\
\midrule
NHI Drug $\rightarrow$ ATC   & 85,245 & 3,566  & \textbf{23.9:1} \\
NHI Order $\rightarrow$ CPT  & 8,117  & 3,262  & 2.5:1 \\
NHI Order $\rightarrow$ LOINC & 2,628 & 1,527  & 1.7:1 \\
ICD-9 $\rightarrow$ ICD-10  & 18,662 & 16,418 & 1.1:1 \\
\midrule
\textbf{Total} & \textbf{114,652} & \textbf{24,773} & \textbf{4.6:1} \\
\bottomrule
\end{tabular}
\end{table}

\begin{prompt}[Prompt used for LLM-based CPT mapping]

\ttfamily
You are a medical coding specialist with expertise in both the US CPT

(Current Procedural Terminology) code set and Taiwan's NHI

(National Health Insurance) order/fee code system.

Taiwan NHI descriptions are often in Chinese or mixed Chinese-English.

Your job is to find the single best-matching 5-digit CPT code

for a given Taiwan NHI code.

Rules:

- Return a standard 5-character CPT code (e.g. "99213").

  Do NOT return 7-digit HCPCS-style codes or internal billing codes.

- If the NHI code is an administrative fee

  (ward fee, consultation fee, registration)

  that has no true CPT analog,

  set cpt\_code = null and confidence = "low".

- Use "high" confidence only when the clinical service maps

  unambiguously to one CPT code.

  Use "medium" when the match is reasonable

  but there is category uncertainty.

  Use "low" otherwise.

- Respond with strict JSON matching the requested schema.
\end{prompt}

\subsection{Medical Code Alignment}
\label{subsec:code_alignment}

When tokenizing each clinical event, we first map source-specific codes into a unified international coding space using four mapping dictionaries in the following order:

\begin{enumerate}
    \item \textbf{NHI Drug $\rightarrow$ ATC}: Taiwan-specific drug billing codes mapped to the international Anatomical Therapeutic Chemical (ATC) classification system.
    
    \item \textbf{NHI Order $\rightarrow$ CPT}: Taiwan procedure/order codes mapped to the US Current Procedural Terminology (CPT) system.
    
    \item \textbf{NHI Order $\rightarrow$ LOINC}: Taiwan laboratory order codes mapped to the Logical Observation Identifiers Names and Codes (LOINC) system.
    
    \item \textbf{ICD-9 $\rightarrow$ ICD-10}: Legacy ICD-9 diagnosis codes mapped to ICD-10.
\end{enumerate}

If a source code exists in the mapping dictionary, the mapped international code is emitted; otherwise, the original local code is retained. This alignment enables the model to share unified tokens for equivalent drugs, procedures, laboratory tests, and diagnoses across all datasets.

This mapping process is executed once during dataset construction rather than during training or inference. The resulting mappings are stored as a frozen JSON dictionary used by the preprocessing pipeline. The LLM is only queried when refreshing or expanding the dictionary.

\paragraph{Step A: Inventory Unmapped Codes.}
We first enumerate all unmapped NHI codes and collect their raw descriptions, English translations, and TF-IDF-based candidate matches into a unified JSON file. This stage includes approximately 3,500 NHI procedure codes for CPT mapping and 1,200 NHI material codes for HCPCS mapping, totaling roughly 4,700 unmapped codes.

\paragraph{Step B: LLM-based Code Mapping.}
We then submit the full set of $\sim$4,700 codes to the Anthropic Batch API in a single batch request. For each code, Claude generates its best CPT or HCPCS match, a self-reported confidence score, and a brief rationale. The full mapping process costs approximately \$11 and completes in around 15 minutes.

\paragraph{Step C: Confidence-based Filtering.}
Finally, mappings labeled as low-confidence or null are discarded, while high- and medium-confidence predictions are retained as the final mapping dictionary. This process yields 3,475 NHI-to-CPT mappings.

We use Claude Haiku 4.5 because the task is structured single-token-system mapping rather than complex reasoning, and Haiku handles Chinese / mixed Chinese-English medical text well in our pre-tests. Detailed prompt is shown above.

In EHRShot, we first leverage the official mapping tables to align SNOMED and RxNorm codes into the unified coding space. For codes without direct mappings, we adopt a text-based semantic matching approach by embedding the textual descriptions of medical codes using Sentence Transformers~\citep{thakur-2020-AugSBERT} and formulating the alignment task as a dense retrieval problem, a strategy widely used for cross-ontology medical concept alignment~\citep{sung2020biomedical,liu2020self,yuan2022coder}. Specifically, codes from each source ontology are matched to the most semantically similar code in the unified coding space based on embedding similarity, using Faiss~\citep{douze2024faiss} as the retrieval backend. We evaluate several embedding encoders, including SFR-Embedding-Mistral~\citep{SFRAIResearch2024}, Qwen3-Embedding-8B~\citep{zhang2025qwen3}, Linq-Embed-Mistral~\citep{choi2024linq}, and multilingual-e5-large-instruct~\citep{wang2024multilinguale5textembeddings}, with similarity thresholds selected from $\{0.7, 0.8, 0.9\}$.

\subsection{Training dataset}
The Taiwanese National Health Insurance Research Database (NHIRD) is one of the largest and most comprehensive de-identified population-wide EHR datasets globally~\citep{hsieh2019taiwan}, covering over 99\% of the Taiwanese population. Maintained by the National Health Insurance Administration in collaboration with the Ministry of Health and Welfare, the dataset contains longitudinal medical records spanning more than two decades, including patient demographics, diagnoses, prescriptions, medical procedures, clinical events, and hospital visits collected from healthcare facilities across Taiwan. The NHIRD consists of both registration records and hospital reimbursement claims submitted under the National Health Insurance (NHI) program.

NHIRD is accessible through a controlled access framework. Research institutions in Taiwan may apply for access by obtaining Institutional Review Board (IRB) approval and complying with data use regulations established by the Ministry of Health and Welfare, while international institutions may access the dataset through formal collaborations with Taiwanese research organizations. We obtained access through a collaboration with a Taiwanese medical school and completed the required IRB review process, including submission of the data schema, research protocol, access procedures, and authorized user list. Our study received IRB approval with US data and exemption because it relied exclusively on fully de-identified secondary data and involved no human subjects. To support reproducibility, we will publicly release the complete preprocessing pipeline and modeling code. In addition, pretrained model checkpoints will be made available to researchers who independently obtain NHIRD access through the same regulatory framework, enabling reproducible downstream research while maintaining strict privacy and compliance standards.

\textbf{Data De-identification and Privacy.} All NHIRD data are de-identified following strict protocols mandated by the National Health Insurance Administration (NHIA). Personally identifiable information (PII), such as names and ID numbers, is removed and irreversibly encrypted using non-public anonymization methods~\citep{lin2018data, mohw}. Our study uses only anonymized demographic variables, e.g., age and gender, for model training. Given the level of anonymization and the coarse granularity of these attributes, the risk of patient re-identification is negligible. Hence, our use of NHIRD data fully complies with privacy standards and poses no ethical or legal concerns regarding patient confidentiality. 

\begin{table*}[t]
\centering

\caption{Model hyperparameters under different parameter sizes with million (m) and billion (b).}
\label{tab:model-architecture}
\begin{tabular}{lcccc}
\toprule
\textbf{Parameters} & \textbf{Num of layers} & \textbf{Dimension} & \textbf{Num of heads} & \textbf{Block size}  \\ \midrule
70m                 & 6                      & 512                & 8                     & 2048                 \\
120m                & 6                      & 768                & 8                     & 2048                \\
160m                & 12                     & 768                & 12                    & 2048                \\
260m                & 12                     & 1024               & 16                    & 2048                \\
350m                & 20                     & 1024               & 16                    & 2048                 \\
410m                & 24                     & 1024               & 16                    & 2048                \\
560m                & 22                     & 1280               & 10                    & 2048                \\
720m                & 20                     & 1536               & 12                    & 2048                \\
1b                  & 16                     & 2048               & 8                     & 2048                \\
1.2b                & 20                     & 2048               & 16                    & 2048                \\
1.4b                & 24                     & 2048               & 16                    & 2048                \\
2.1b                & 24                     & 2560               & 16                    & 2048                \\
2.8b                & 32                     & 2560               & 32                    & 2048               \\ 
\bottomrule
\end{tabular}
\end{table*}
\begin{table*}[t]

  \centering
  \caption{Hyperparameters Configurations for EHR-FM}
  \label{tab:hyper}
\begin{tabular}{lcc}
\toprule
\textbf{Hyperparameter} & \textbf{Pretraining} & \textbf{Supervised Fine-tuning} \\
\midrule
Learning Rate & \multicolumn{2}{c}{6e-6 for model size $\ge$ 1B; else 1e-5} \\
Optimizer &\multicolumn{2}{c}{AdamW} \\
Adam $\epsilon$ & \multicolumn{2}{c}{1e-8} \\
Adam Betas ($\beta_1$, $\beta_2$) & \multicolumn{2}{c}{(0.9, 0.999)} \\
Weight decay & \multicolumn{2}{c}{0.01} \\
Gradient Norm & \multicolumn{2}{c}{0.1} \\
Scheduler & \multicolumn{2}{c}{Warmup-Stable-Decay} \\
Warmup Ratio &\multicolumn{2}{c}{0.1} \\
Stable Ratio & \multicolumn{2}{c}{0.8} \\
Decay Ratio&\multicolumn{2}{c}{0.1}  \\
Batch Size & 64 & 128 \\
Epochs & -& 5 \\

\bottomrule
\end{tabular}
\end{table*}
\section{Implementation Details} 
\label{sec:implement}
\textbf{Tokenization and inputs.} 
We map all medical codes to unique indices ranging from 0 to the total number of unique medical codes, with a token vocabulary size of 225,568, including demographic, time, and special tokens. Since all tokens are treated as atomic units, no additional tokenization is required. The input sequence length is limited to 2,048 tokens. For patient records with EHR sequences exceeding this limit, we avoid truncation during pretraining and instead split the sequences into non-overlapping chunks, processing them across multiple training steps. However, during fine-tuning, inputs longer than 2,048 tokens are truncated from starting, as they must be processed as single sequences.

\textbf{Backbone architecture.} We use the Pythia architecture as the backbone of \method. Pythia is a family of decoder-only autoregressive language models, ranging from 70m to 12b parameters, designed for scalable and consistent research. Its feedforward architecture features rotary embeddings for positional encoding, untied embedding layers, and parallelized flash attention for efficient training. Table~\ref{tab:model-architecture} outlines the model architecture, while Table~\ref{tab:hyper} details the hyperparameters for pretraining and supervised fine-tuning. All models are trained on 8 A100-SXM4-40GB GPUs.

\textbf{Baselines.} For the tree-based baselines, XGBoost and LightGBM, we utilize their official Python packages, xgboost~\footnote{https://xgboost.readthedocs.io/en/stable/index.html} and lightgbm~\footnote{https://lightgbm.readthedocs.io/en/stable/}, respectively. For language model baselines, we convert medical codes into text using language model vocabularies and concatenate them into sequences representing patient histories for fine-tuning. All codes are mapped to standardized textual descriptions using official code mapping tables. Each medical code, representing a diagnosis, surgery, treatment, or medication, is transformed accordingly. Patient histories are then constructed by concatenating the medical text, similar to processing text documents in language models. The sequence length limit is 1024 for BioMedLM and 2048 for Qwen for computation efficiency .

\textbf{Training Configurations.} 
Table \ref{tab:compute-optimal-flops} shows the required training steps and total tokens for \method~at various FLOP targets. We adopt this configuration to investigate the scaling laws for \method, aiming to determine the optimal model and data scales under a fixed FLOP budget. Their corresponding FLOPs, number of parameters, and GPU training hours are listed in Table~\ref{tab:gpu_hours}. We calculate the FLOPs with Pytorch built-in flops counter.
\begin{table*}[t]
\centering
\caption{Estimated FLOPs requirement for \method. The required training steps and total tokens (inside the parentheses) for different model sizes are reported here.}
\label{tab:compute-optimal-flops}
\resizebox{1.0\textwidth}{!}{\begin{tabular}{lllllllll}
\toprule
\textbf{FLOPS} & \textbf{1e+18}   & \textbf{4e+18}   & \textbf{8e+18}   & \textbf{2e+19}    & \textbf{4e+19}    & \textbf{8e+19}    & \textbf{1e+20}     & \textbf{2e+20}     \\ \midrule
\method-70m     & 10071 (1.32e+09) & 40287 (5.28e+09) & 80575 (1.06e+10) & 201439 (2.64e+10) & 402878 (5.28e+10) & 805757 (1.06e+11) & 1007197 (1.32e+11) & 2014394 (2.64e+11) \\
\method-120m    & 6247 (8.19e+08)  & 24990 (3.28e+09) & 49981 (6.55e+09) & 124952 (1.64e+10) & 249905 (3.28e+10) & 499810 (6.55e+10) & 624763 (8.19e+10)  & 1249526 (1.64e+11) \\
\method-160m    & 4800 (6.29e+08)  & 19202 (2.52e+09) & 38405 (5.03e+09) & 96014 (1.26e+10)  & 192029 (2.52e+10) & 384058 (5.03e+10) & 480072 (6.29e+10)  & 960145 (1.26e+11)  \\
\method-260m    & 3252 (4.26e+08)  & 13011 (1.71e+09) & 26022 (3.41e+09) & 65057 (8.53e+09)  & 130114 (1.71e+10) & 260228 (3.41e+10) & 325285 (4.26e+10)  & 650570 (8.53e+10)  \\
\method-350m    & 2421 (3.17e+08)  & 9685 (1.27e+09)  & 19371 (2.54e+09) & 48429 (6.35e+09)  & 96858 (1.27e+10)  & 193716 (2.54e+10) & 242145 (3.17e+10)  & 484290 (6.35e+10)  \\
\method-410m    & 2147 (2.81e+08)  & 8588 (1.13e+09)  & 17176 (2.25e+09) & 42941 (5.63e+09)  & 85882 (1.13e+10)  & 171765 (2.25e+10) & 214706 (2.81e+10)  & 429412 (5.63e+10)  \\
\method-720m    & 1302 (1.71e+08)  & 5209 (6.83e+08)  & 10418 (1.37e+09) & 26045 (3.41e+09)  & 52090 (6.83e+09)  & 104180 (1.37e+10) & 130225 (1.71e+10)  & 260451 (3.41e+10)  \\
\method-1b      & 964 (1.26e+08)   & 3857 (5.06e+08)  & 7714 (1.01e+09)  & 19285 (2.53e+09)  & 38570 (5.06e+09)  & 77141 (1.01e+10)  & 96426 (1.26e+10)   & 192853 (2.53e+10)  \\
\method-1.2b    & 818 (1.07e+08)   & 3273 (4.29e+08)  & 6547 (8.58e+08)  & 16369 (2.15e+09)  & 32739 (4.29e+09)  & 65478 (8.58e+09)  & 81848 (1.07e+10)   & 163696 (2.15e+10)  \\
\method-1.4b    & 710 (9.32e+07)   & 2843 (3.73e+08)  & 5687 (7.46e+08)  & 14219 (1.86e+09)  & 28439 (3.73e+09)  & 56879 (7.46e+09)  & 71098 (9.32e+09)   & 142197 (1.86e+10)  \\
\method-2.1b    & 486 (6.38e+07)   & 1946 (2.55e+08)  & 3892 (5.10e+08)  & 9732 (1.28e+09)   & 19464 (2.55e+09)  & 38929 (5.10e+09)  & 48662 (6.38e+09)   & 97324 (1.28e+10)   \\
\method-2.8b    & 382 (5.01e+07)   & 1529 (2.00e+08)  & 3058 (4.01e+08)  & 7646 (1.00e+09)   & 15292 (2.00e+09)  & 30584 (4.01e+09)  & 38230 (5.01e+09)   & 76460 (1.00e+10)   \\ 
\bottomrule
\end{tabular}}
\end{table*}

\section{Usage of Large Language Model}
We used large language models to assist in polishing the writing of this paper. Specifically, LLMs were employed to correct grammar, paraphrase sentences, and improve readability and flow. The scientific ideas, experiments, and analyses were fully conducted by the authors, with LLM use limited to enhancing clarity and smoothness of expression.

\section*{NeurIPS Paper Checklist}

The checklist is designed to encourage best practices for responsible machine learning research, addressing issues of reproducibility, transparency, research ethics, and societal impact. Do not remove the checklist: {\bf The papers not including the checklist will be desk rejected.} The checklist should follow the references and follow the (optional) supplemental material.  The checklist does NOT count towards the page
limit. 

Please read the checklist guidelines carefully for information on how to answer these questions. For each question in the checklist:
\begin{itemize}
    \item You should answer \answerYes{}, \answerNo{}, or \answerNA{}.
    \item \answerNA{} means either that the question is Not Applicable for that particular paper or the relevant information is Not Available.
    \item Please provide a short (1--2 sentence) justification right after your answer (even for \answerNA). 
\end{itemize}

{\bf The checklist answers are an integral part of your paper submission.} They are visible to the reviewers, area chairs, senior area chairs, and ethics reviewers. You will also be asked to include it (after eventual revisions) with the final version of your paper, and its final version will be published with the paper.

The reviewers of your paper will be asked to use the checklist as one of the factors in their evaluation. While \answerYes{} is generally preferable to \answerNo{}, it is perfectly acceptable to answer \answerNo{} provided a proper justification is given (e.g., error bars are not reported because it would be too computationally expensive'' or ``we were unable to find the license for the dataset we used''). In general, answering \answerNo{} or \answerNA{} is not grounds for rejection. While the questions are phrased in a binary way, we acknowledge that the true answer is often more nuanced, so please just use your best judgment and write a justification to elaborate. All supporting evidence can appear either in the main paper or the supplemental material, provided in appendix. If you answer \answerYes{} to a question, in the justification please point to the section(s) where related material for the question can be found.

IMPORTANT, please:
\begin{itemize}
    \item {\bf Delete this instruction block, but keep the section heading ``NeurIPS Paper Checklist"},
    \item  {\bf Keep the checklist subsection headings, questions/answers and guidelines below.}
    \item {\bf Do not modify the questions and only use the provided macros for your answers}.
\end{itemize}


\begin{enumerate}

\item {\bf Claims}
    \item[] Question: Do the main claims made in the abstract and introduction accurately reflect the paper's contributions and scope?
    \item[] Answer: \answerYes{} 
    \item[] Justification:Our experiment in Section~\ref{sec:scaling_laws} and Section~\ref{sec:overall_results} shows the scaling law behavior compatible with general LLMs and performance boost on harder cancer prediction tasks by scaling, as well as confirms our accurate cancer pre-screening using only historical EHR data. In Section~\ref{sec:generalization}, we show that our model outperforms all state-of-the-art baselines on EHR datasets that come from completely different healthcare systems, which validates our claim to generalize to other healthcare systems.

\item {\bf Limitations}
    \item[] Question: Does the paper discuss the limitations of the work performed by the authors?
    \item[] Answer: \answerYes{} 
    \item[] Justification: Refer to Section~\ref{sec:limitation} for our discussion on limitations of our work.


\item {\bf Theory assumptions and proofs}
    \item[] Question: For each theoretical result, does the paper provide the full set of assumptions and a complete (and correct) proof?
    \item[] Answer: \answerNA{} 
    \item[] Justification: 

    \item {\bf Experimental result reproducibility}
    \item[] Question: Does the paper fully disclose all the information needed to reproduce the main experimental results of the paper to the extent that it affects the main claims and/or conclusions of the paper (regardless of whether the code and data are provided or not)?
    \item[] Answer: \answerYes{} 
    \item[] Justification: We provide a very detailed explanation and discussion of our experiments and additional contents for a dataset in the Appendix.

\item {\bf Open access to data and code}
    \item[] Question: Does the paper provide open access to the data and code, with sufficient instructions to faithfully reproduce the main experimental results, as described in supplemental material?
    \item[] Answer: \answerNo{} 
    \item[] Justification: We provide open access to the source code; however, we are unable to release our healthcare data due to strict patient privacy regulations.

\item {\bf Experimental setting/details}
    \item[] Question: Does the paper specify all the training and test details (e.g., data splits, hyperparameters, how they were chosen, type of optimizer) necessary to understand the results?
    \item[] Answer:\answerYes{} 
    \item[] Justification: We provide a detailed dataset curation process in Section~\ref{sec:data_curation} and data split information in Section~\ref{sec:experiment-setup}.

\item {\bf Experiment statistical significance}
    \item[] Question: Does the paper report error bars suitably and correctly defined or other appropriate information about the statistical significance of the experiments?
    \item[] Answer: \answerYes{} 
    \item[] Justification: EHRShot results are averaged over five random seeds, and our model outperforms the baselines with statistical significance.

\item {\bf Experiments compute resources}
    \item[] Question: For each experiment, does the paper provide sufficient information on the computer resources (type of compute workers, memory, time of execution) needed to reproduce the experiments?
    \item[] Answer:  \answerYes{} 
    \item[] Justification: We provide software and hardware configuration for developing and training in the Appendix.
    
\item {\bf Code of ethics}
    \item[] Question: Does the research conducted in the paper conform, in every respect, with the NeurIPS Code of Ethics \url{https://neurips.cc/public/EthicsGuidelines}?
    \item[] Answer: \answerYes{} 
    \item[] Justification: We confirm our research meets the NeurIPS Code of Ethics. It uses de-identified, IRB-exempt EHR data, without human subjects and ensures privacy. The dataset access is under the formal research agreement with collaborated institutions. We thoroughly benchmark our model considering generalization, fairness, and bias in our experiments and appendix. All code for data processing, model training, and evaluation will be released under the MIT license.

\item {\bf Broader impacts}
    \item[] Question: Does the paper discuss both potential positive societal impacts and negative societal impacts of the work performed?
    \item[] Answer:  \answerYes{} 
    \item[] Justification: Refer Section~\ref{sec:impact_statement} for our impact statement and discussion.
    
\item {\bf Safeguards}
    \item[] Question: Does the paper describe safeguards that have been put in place for responsible release of data or models that have a high risk for misuse (e.g., pre-trained language models, image generators, or scraped datasets)?
    \item[] Answer: \answerYes{} 
    \item[] Justification: We have stated that data access complies with regulations from governing agencies, and we only release model weights to research parties with the same data access licenses. This restriction serves as a safeguard to prevent misuse, ensuring that models trained on sensitive EHR data are only used by authorized researchers under appropriate ethical and legal oversight.

\item {\bf Licenses for existing assets}
    \item[] Question: Are the creators or original owners of assets (e.g., code, data, models), used in the paper, properly credited and are the license and terms of use explicitly mentioned and properly respected?
    \item[] Answer: \answerYes{} 
    \item[] Justification: We have properly cited the original sources for all datasets and our pretraining EHR dataset. We have also started the data access process and ensured that all licenses and usage terms are explicitly acknowledged and followed.

\item {\bf New assets}
    \item[] Question: Are new assets introduced in the paper well documented and is the documentation provided alongside the assets?
    \item[] Answer: \answerYes{} 
    \item[] Justification: We have a detailed and complete introduction to our curated dataset from our source EHR. We also clearly describe our model configuration, architecture, and codes as well as data processing pipeline which will be released under MIT license.

\item {\bf Crowdsourcing and research with human subjects}
    \item[] Question: For crowdsourcing experiments and research with human subjects, does the paper include the full text of instructions given to participants and screenshots, if applicable, as well as details about compensation (if any)? 
    \item[] Answer: \answerNA{} 
    \item[] Justification: 

\item {\bf Institutional review board (IRB) approvals or equivalent for research with human subjects}
    \item[] Question: Does the paper describe potential risks incurred by study participants, whether such risks were disclosed to the subjects, and whether Institutional Review Board (IRB) approvals (or an equivalent approval/review based on the requirements of your country or institution) were obtained?
    \item[] Answer:\answerYes{} 
    \item[] Justification: We have described the procedure to get IRB as well as how it ends up with exemption in the appendix.

\item {\bf Declaration of LLM usage}
    \item[] Question: Does the paper describe the usage of LLMs if it is an important, original, or non-standard component of the core methods in this research? Note that if the LLM is used only for writing, editing, or formatting purposes and does \emph{not} impact the core methodology, scientific rigor, or originality of the research, declaration is not required.
    \item[] Answer:\answerNA{} 
    \item[] Justification:

\end{enumerate}

\end{document}